# SCRIBBLE-SUPERVISED TARGET EXTRACTION METHOD BASED ON INNER STRUCTURE-CONSTRAINT FOR REMOTE SENSING IMAGES

*Yitong Li[1], Chang Liu[1], Jie Ma[1*]*

[1]Beijing Foreign Studies University

## ABSTRACT

Weakly supervised learning based on scribble annotations in target extraction of remote sensing images has drawn much interest due to scribbles' flexibility in denoting winding objects and low cost of manually labeling. However, scribbles are too sparse to identify object structure and detailed information, bringing great challenges in target localization and boundary description. To alleviate these problems, in this paper, we construct two inner structure-constraints, a deformation consistency loss and a trainable active contour loss, together with a scribble-constraint to supervise the optimization of the encoder-decoder network without introducing any auxiliary module or extra operation based on prior cues. Comprehensive experiments demonstrate our method's superiority over five state-of-the-art algorithms in this field. Source code is available at https://github.com/yitongli123/ISC-TE.

## 1. INTRODUCTION

Target extraction [1, 2] in the field of remote sensing aims at segmenting the target regions in a remote sensing image (RSI) through generating a binary pixel-wise mask. Recently, weakly-supervised target extraction methods based on deep learning [2] have been widely explored in academic community owing to the advantage of saving labor and time consumption in labeling. Weak annotations include image-level tags, bounding-box labels and scribbles, where scribbles are made up of three kinds of pixels with different grey values to signify target region, background and unknown area respectively. Among them, scribble annotation has drawn much interest for its flexibility in denoting winding objects and low cost of manually labeling.

However, scribbles are too sparse to identify object structure and detailed information, causing inferior performance in boundary localization and background suppression compared with fully-supervised methods. To seek for competitive results, previous works came up with various skills. Scribble2Label [3] combined pseudo-labeling and label filtering to carry iterative annotation expansion and improve scribble credibility. ScribbleSaliency [4] employed a well-trained external algorithm in the additional edge detection network to introduce prior supervision information, and exploited a scribble boosting scheme that relies on external segmentation approach DenseCRF [5]. RD-DN [2] designed a dense network combing residual skip connection with hybrid dilated convolution and a progressive label updating strategy using morphological dilation operation. On the basis of the auto-encoder structure, SCWSSOD [6] proposed a local coherence loss for scribble propagation, a saliency structure consistency loss for ensuring consistent results with multi-sized same image as input and an aggregation module for integrating multi-level features. Apart from a main boundary-aware saliency prediction network using auto-encoder and dense aggregation, SBA-Net [7] designed a boundary label generation module (BLG) working by making class activation prediction close to the image-level tags for generating high-confidence boundary labels to supervise boundary-aware module (BAM) to learn credible boundary semantics from low-level features and input image.

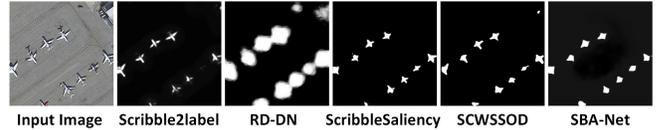

**Fig 1.** Extraction masks of five scribble-supervised SOTAs.

With reference to above analysis, previous methods mostly construct complicated network or add external algorithms to increase supervision data or empirical information, which brings numerous parameters and time-consuming training. Meanwhile, RSI-oriented target extraction faces greater challenges considering RSIs' inconsistency inside the target regions and disordered distribution outside. As a result, they fail to strike the trade-off between visual performance and computing efficiency, which can be embodied in two aspects: poor boundary localization and unideal background suppression. As shown in Fig 1, we carry target extraction with airplane as target using five state-of-art scribble-supervised models. The airplanes' boundaries in these masks are too coarse to be consistent with the actual edges in remote sensing images. Moreover, they will easily make errors under the disturbance of object shadow, boarding bridge and so on.

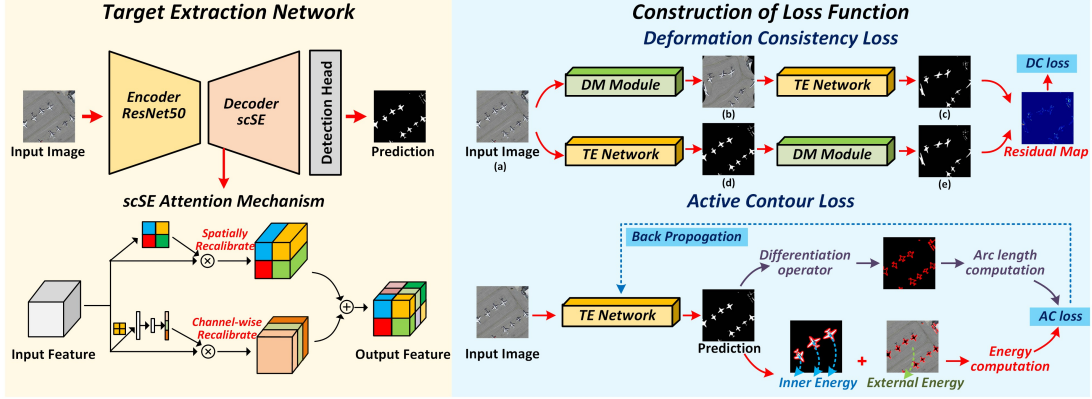

**Fig 2.** The overview of our network structure and constraint construction.

In this paper, we propose a novel scribble-supervised method of target extraction based on inner structure-constraint for RSIs. Exploiting a simple end-to-end network of encoder-decoder structure, we construct two inner structure-constraints to distill input image's intrinsic features and generate extraction masks with accurate boundaries and clear background, without introducing any auxiliary module or extra operation based on prior cues. As shown in Fig 2, on the one hand, to discriminate target regions with integral object structure and detailed edge, we design a deformation consistency loss according to the idea that the order of extraction operator and deformation operator can be exchanged. On the other hand, to reach better performance under challenging scenarios where many deceptive factors exist in background, the Eulerian energy function in active contour model (ACM) [8] is served as an active contour loss, rather than the postprocessing operation, to automatically capture target boundaries with fine-grained details and high accuracy. Comprehensive experiments show that our method has achieved a better performance than five state-of-art scribble-supervised algorithms.

## 2. METHODOLOGY

In our proposed network, we utilize U-Net [9] with ResNet-50 [10] encoder as our framework. Besides, for the purpose of discriminately integrating spatial information in different channels, we apply the spatial and channel Squeeze & Excitation (scSE) attention mechanism [11] in decoder to stimulate human perception. The decoder output is a two-channel possibility prediction for target and background with values in range of 0 to 1. As for inner structure-constraints, detailed descriptions are presented as follows.

### 2.1. Deformation Consistency Constraint

For scribble-supervised methods, the main problem lies in that scribbles provide scanty information for object structure and boundary details. To alleviate it, we choose to intensify the training objective with a constraint based on visual tasks. For an ideal target extraction model, the order of extracting operation and deforming operation should be exchangeable. In other words, illustrated in Fig 2, the model prediction (c) for a deformed input image (b) should be consistent with the result (e) of deforming the model prediction (d) for the input image's undeformed version (a), where two deforming operations share the same hyper-parameters and procedures. Applying L1 penalty on residual map of the two masks (c and e) to encourage them to be similar, our deformation consistency constraint is defined as:

$$L_{dc} = \left\| D\big(F(I;\theta);\alpha,\beta\big) - F\big(D(I;\alpha,\beta);\theta\big) \right\|_1, \quad (1)$$

where $I$ is the input image, $F(\cdot;\theta)$ denotes our proposed model with trainable network parameters $\theta$, $D(\cdot;\alpha,\beta)$ denotes the deforming operation with two hyper-parameters $\alpha$ and $\beta$. It is worth noting that the deformation operator can be directly conducted without training.

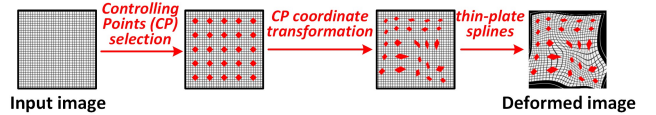

**Fig 3.** Procedures of the deforming operation.

As shown in Fig 3, the deformation here consists of three steps: selecting controlling points (CP), CP coordinate transformation and thin-plate splines (TPS) [12]. Introducing the hyper-parameter $\alpha$ as coordinate interval for CP, we decide $n \times n$ controlling points in the grided image ranging $[-1,1] \times [-1,1]$ with uniform spacing, where $n = \lfloor 2/\alpha \rfloor + 1$. Then we get a CP coordinate matrix $C \in \mathbb{R}^{N \times 2}$:

$$C = \begin{bmatrix} C_1^x & C_2^x & \cdots & C_N^x \\ C_1^y & C_2^y & \cdots & C_N^y \end{bmatrix}^T, N = n^2, \quad (2)$$

For CP coordinate transformation, we introduce the hyper-parameter $\beta$ to generate coordinate variation matrix $V \in \mathbb{R}^{N \times 2}$ which adjusts twisting degree. Every value in $V$ is randomly sampled in the uniform distribution of $[-\beta, \beta]$. Thus, the CP coordinate matrix after deformation can be computed by $\overline{C} = C + V$.

To produce the final deformed image, we apply the interpolation method of TPS [12] on other pixels except the controlling points for obtaining their coordinates after deformation. Given $C$, $\overline{C}$ and input image, TPS [12] can find the optimal mapping matrix under the condition of minimum energy, thereby creating a new deformed image.

## 2.2. Trainable Active Contour Constraint

Considering that some high-frequency data including edge, fine structure and other regions with drastic changes are prone to be lost after repeated down-sampling in convolutional neural network (CNN), we exploit the traditional ACM [8] algorithm's advantage of curve evolution and edge positioning in CNN to help our model overcome information loss. Based on the rule of minimizing Eulerian energy function in ACM [8], we design a trainable active contour constraint for better background suppression and boundary description, which consists of two constraint items: boundary curve length and image coherence.

$$L_{ac} = Length + \lambda L_{ic}, \qquad (3)$$

where the first item calculates the arc-length of boundary for finding the most fitting one which can encircle all of pixels in target region with the shortest curve, $\lambda$ is a hyper-parameter controlling the two items' constraining proportion.

Specifically, we compute the arc-length on the predicted extraction mask. Defining $u$ as prediction value ranging in [0,1], $\nabla u_{i,j}^d (d \in \{x, y\})$ indicates the partial differentiation at pixel $(i,j)$ of the mask on $\bar{x}$ or $\bar{y}$ direction. It is worth noting that although $U$ refers to the entire region of the mask, pixels in target region and background have zero differentiation. Thus, only pixels on the boundary are counted in. To avoid calculating the square root of zero, we set $\varepsilon = 1 \times e^{-8}$.

$$Length = \sum_{U}^{i,j} \sqrt{\left|\left(\nabla u_{i,j}^x\right)^2 + \left(\nabla u_{i,j}^y\right)^2\right| + \varepsilon}, \qquad (4)$$

For image coherence, the second item measures internal dispersion degree of two subregion distributions. The two subregions refer to predicted target region and predicted background, which are separated by predicted boundary curve. By minimizing this item, we can guarantee the data stability and coherence both inside and outside the predicted target region, which accords with the common sense that edges locate at places with high-frequency changes while non-edge region is relatively smoother.

$$L_{ic} = \int_{i \in T} (\sigma_t - p_i)^2 di + \int_{i \in B} (\sigma_b - p_i)^2 di, \qquad (5)$$

where $p_i$ is the pixel value with position index $i$ in input image, $T$ and $B$ denote the position index collections of pixels in predicted target region and background respectively, $\sigma_t$ and $\sigma_b$ are the mean values calculated in predicted target region and background respectively.

## 2.3. Overall Objective Function

The overall objective function is composed of three parts. Besides of the above two structure-constraints, a scribble-constraint is computed with scribble annotations using partial cross entropy loss:

$$L_{pc} = \sum_{i \in S} -y_i log(y_i^p) - (1 - y_i) log(1 - y_i^p), \qquad (6)$$

where $S$ is the position index collection of annotated pixels in scribbles, $y_i$ is the ground-truth value given by scribbles for position $i$, $y_i^p$ is the predicted value for position $i$.

Finally, our overall objective function can be written as:
$$L_{total} = L_{dc} + L_{ac} + L_{pc}. \qquad (7)$$

## 3. EXPERIMENTS AND DISCUSSION

### 3.1. Data Description and Implementation Details

*Data Description* We evaluate the proposed framework on airport satellite images from Google with three channels and a resolution of 0.5m to 1m. The airplanes are selected as targets in our work. In the scribble annotations, target regions and background are labeled by straight lines with pixel values of 1 and 0 respectively. We adopt 394 raw images with size of 768 × 768 × 3 for training and 12 raw images with size of 768 × 768 × 3 for testing. In training phase, we first randomly crop input images with size of 256 × 256 × 3 before feeding them into the model.

*Implementation Details* We train our network on one NVIDIA RTX 3090 GPU with PyTorch framework. Inspired by Scribble2Label [3], we first train our network with the overall objective function for 500 epochs in warming-up stage, then a label-constraint computed on un-scribbled pixels in the updating filtered pseudo-label is added with a weight as 0.5 to the whole loss function for the next 500 epochs. Both stages are trained with RAdam optimizer, where the initial learning rate is $3 \times e^{-4}$ and the learning rate decay is $5 \times e^{-5}$. As for hyper-parameters, $\alpha = 0.3$, $\beta = 0.7$, and the training batch size is 16.

### 3.2. Comparisons with State-of-the-Art Methods

As shown in Fig 4, we visually compare our model with five scribbled-supervised SOTAs in target extraction task, including Scribble2Label (2020) [3], ScribbleSaliency (2020) [4], RD-DN (2021) [2], SCWSSOD (2021) [6] and SBA-Net (2022) [7]. Quantitatively, we compare these methods on the commonly used evaluation metrics Precision, Recall and F-measure in Table 1.

It can be observed that targets detected by Scribble2Label [3] are incoherent and incomplete with coarse boundaries owing to the lack of structure-oriented component or loss function. Although ScribbleSaliency [4], SCWSSOD [6] and SBA-Net [7] are equipped with auxiliary module or constraint focusing on inner consistency and edge enhancement, they fail to extract integral targets and accurate boundaries, which is also evidenced by their low Recall and F-measure scores. RD-DN [2] can achieve the highest Recall, however, its Precision and F-measure scores reach the lowest, which can be explained by its large-scale target prediction with many erroneous pixels and poor

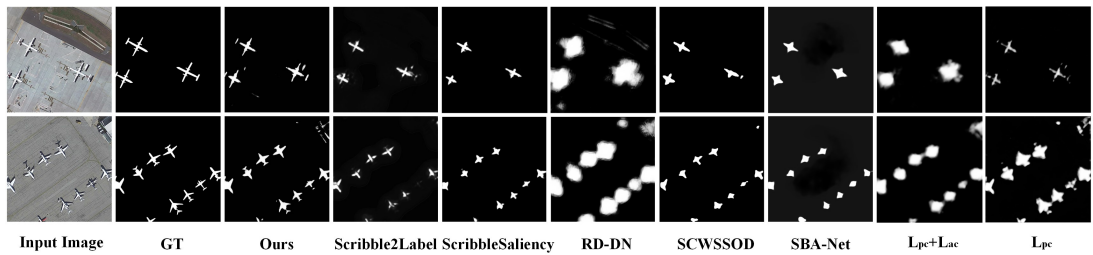

**Input Image · GT · Ours · Scribble2Label · ScribbleSaliency · RD-DN · SCWSSOD · SBA-Net · $L_{pc}+L_{ac}$ · $L_{pc}$**

**Fig 4.** Qualitative comparison of the proposed model, five state-of-the-art methods and ablation study results.

**Table 1.** The Evaluation Metrics Result

| Method | Precision | Recall | F-measure |
|---|---|---|---|
| Scribble2Label | 92.1% | 51.5% | 76.3% |
| ScribbleSaliency | 91.3% | 39.9% | 67.3% |
| RD-DN | 20.6% | **98.7%** | 25.0% |
| SCWSSOD | 85.5% | 49.6% | 72.6% |
| SBA-Net | 82.6% | 40.9% | 64.8% |
| **Ours** | **92.2%** | 55.3% | **77.6%** |

resistance to the distraction of background factors since the ignorance of structure-aware supervision. In contrast, our model attains superior performance by capturing targets with sharp boundaries, spatial smoothness and visual wholeness, as well as the highest Precision and F-measure. There is a flaw, though, it is slightly prone to be deceived by target-like background objects due to the ACM-based loss.

### 3.3. Ablation Study

***W/O*** $L_{dc}$ To evaluate the effectiveness of the deformation consistency constraint, we train the proposed network only with $L_{pc}$ and $L_{ac}$. Ambiguous outline and redundant areas of predicted targets can be seen in the 9th column of Fig 4, proving the precise boundary localization function of $L_{dc}$.

***W/O*** $L_{dc}+L_{ac}$ In the 10th column of Fig 4, we demonstrate the results of training with $L_{pc}$ alone. Comparing with the full model, there are problems of incoherent internal structure and many false positives, confirming the improvement in structure detection and target identification by adding the trainable active contour constraint.

### 4. CONCLUSION

In this paper, we propose a novel scribble-based weakly-supervised method of target extraction based on inner structure-constraint for remote sensing images. To increase supervision information, we construct two inner structure-constraints to distill input image's intrinsic features and generate extraction masks with accurate boundaries and clear background. The deformation consistency constraint is designed according to the exchangeability between the two visual tasks of target extraction and deformation, while the active contour constraint is applied on the basis of the curve evolution algorithm ACM. Both qualitative comparison and quantitative evaluation metrics show the superiority of our method with sparse scribble annotations.

### ACKNOWLEDGEMENT

This work is supported by the National Natural Science Foundation of China (62101052).